\documentclass{ifacconf}

\usepackage{graphicx}      
\usepackage{natbib}        
\usepackage{macros}
\usepackage{epsfig} 
\usepackage{graphics}
\usepackage{latexsym} 
\usepackage{psfrag}
\usepackage{amsmath}
\newtheorem{theorem}{Theorem}
\usepackage{amssymb}
\usepackage{multicol}
\usepackage{url}
\usepackage{bm} 
\usepackage{color} 
\let\oldref\ref
\renewcommand{\ref}[1]{(\oldref{#1})}
\usepackage{subcaption}
\usepackage{refcount} 
\usepackage{array}

\usepackage[absolute,overlay]{textpos}

\newcolumntype{C}[1]{>{\centering\arraybackslash}m{#1}}

\usepackage{color}
\usepackage{amsmath}
\usepackage{amssymb,latexsym}
\usepackage[colorinlistoftodos, color=blue!20]{todonotes}

\definecolor{amaranth}{rgb}{0.9, 0.17, 0.31}
\definecolor{burgundy}{rgb}{0.5, 0.0, 0.13}
\definecolor{forestgreen(traditional)}{rgb}{0.0, 0.27, 0.13}
\definecolor{ultramarine}{rgb}{0.07, 0.04, 0.56}
\definecolor{royalfuchsia}{rgb}{0.79, 0.17, 0.57}
\def\SOthree{\mathbb{SO}(3)}
\def\sothree{\mathfrak{so(3)}}

\def\skewsym #1#2#3 {\left[ \begin{array}{ccc} 0 & -#3 & #2 \\
#3 & 0  & -#1   \\
-#2 &  #1 &  0 \\ \end{array} \right]}

\def\bei{\begin{itemize}}
\def\ei{\end{itemize}}
\def\been{\begin{enumerate}}
\def\een{\end{enumerate}}

\usepackage{tikz}

\newcommand\submittedtext{%
  \footnotesize This work has been submitted to IFAC for possible publication.}

\newcommand\submittednotice{%
\begin{tikzpicture}[remember picture,overlay]
\node[anchor=south,yshift=70pt] at (current page.south) {\parbox{\dimexpr1\textwidth-\fboxsep-\fboxrule\relax}{\submittedtext}};
\end{tikzpicture}%
}
\begin{document}
\begin{frontmatter}

\title{Haptic-based Complementary Filter for Rigid Body Rotations} 


\author[First]{Amit Kumar\textsuperscript{1}}
\author[First]{Domenico Campolo\textsuperscript{1}} 
\author[Second]{Ravi N. Banavar\textsuperscript{2}}
\newline
\address[First]{\textsuperscript{1}School of Mechanical and Aerospace Engineering, Nanyang Technological University, Singapore, Singapore. (e-mail: amit011@e.ntu.edu.sg, d.campolo@ntu.edu.sg).}
\address[Second]{\textsuperscript{2}Centre for Systems and Control, Indian Institute of Technology Bombay, Mumbai, India (e-mail: banavar@iitb.ac.in)}

\begin{abstract}                
The non-commutative nature of 3D rotations poses well-known challenges in generalizing planar problems to three-dimensional ones, even more so in contact-rich tasks where haptic information (i.e., forces/torques) is involved. In this sense, not all learning-based algorithms that are currently available generalize to 3D orientation estimation. Non-linear filters defined on the special orthogonal group, $\mathbf{\SOthree}$, are widely used with inertial measurement sensors; however, none of them have been used with haptic measurements. This paper presents a unique complementary filtering framework that initially interprets the geometric shape of objects in the form of superquadrics, exploits the symmetry of $\mathbf{\SOthree}$, and uses force and vision sensors as measurements to provide an estimate of orientation. The framework's robustness and almost global stability are substantiated by a set of experiments on a dual-arm robotic setup.
\end{abstract}

\begin{keyword}
Nonlinear observers and filters, haptics, dual-arm robotic manipulation, complementary filter, special orthogonal group
\end{keyword}

\end{frontmatter}

\submittednotice

\section{Introduction}
With the recent advancements in robotic manipulation, a big question remains: `If humans can perform various manipulation tasks dexterously, why can't we program robots to do the same?.' Studies from neurobiology have proven that people with impaired tactile sensibility find it challenging to control object manipulations~(\citet{JohanssonTactile}). It is then imperative to include haptic information (sense of touch and motion) for robotic operations such as those performed by humanoids or robotic arms~(\citet{dual_arm_Harsha_Prof_Domenico}). This haptic information aids in developing accurate controllers and observers, e.g., for manufacturing and assembly tasks~(\citet{Lin_Prof_Domenico_Insertion_Task}).

Recent literature demonstrates that learning-based methods, such as neural networks, are effective for some vision-based contact-rich manipulation tasks~(\citet{DecomImiLear}). Factors like lighting conditions, distractor objects, backgrounds, table textures, and camera positions play a role in generalizing such methods. Researchers in~(\citet{SimPLE}) have created a manipulation pipeline, SimPLE, for task-aware picking, visuotactile object pose estimation, and motion planning. Given an object's computer-aided design (CAD), it learns to pick and place an object using simulation. The visuotactile object pose estimation combines tactile images and a depth camera to update the estimate of the distribution of possible grasped object poses. Although it achieved 90\% successful placements for six objects that it was tested on, authors mention a known limitation of this pipeline is that it works in open-loop. Once a visuotactile pose estimate is found, it does not update its belief of the object pose. In the traditional Finite Element Analysis (FEA) approach, an accurate simulation may take tens of seconds~(\citet{narang2022factoryfastcontactrobotic}). Additionally, these algorithms are data-hungry and may not be practical in many time-sensitive settings. This necessitates filter designs for haptic feedback that aggregate multiple sensor readings, and reduce failures.

\begin{figure}[!t]
    \includegraphics[width=1\linewidth]{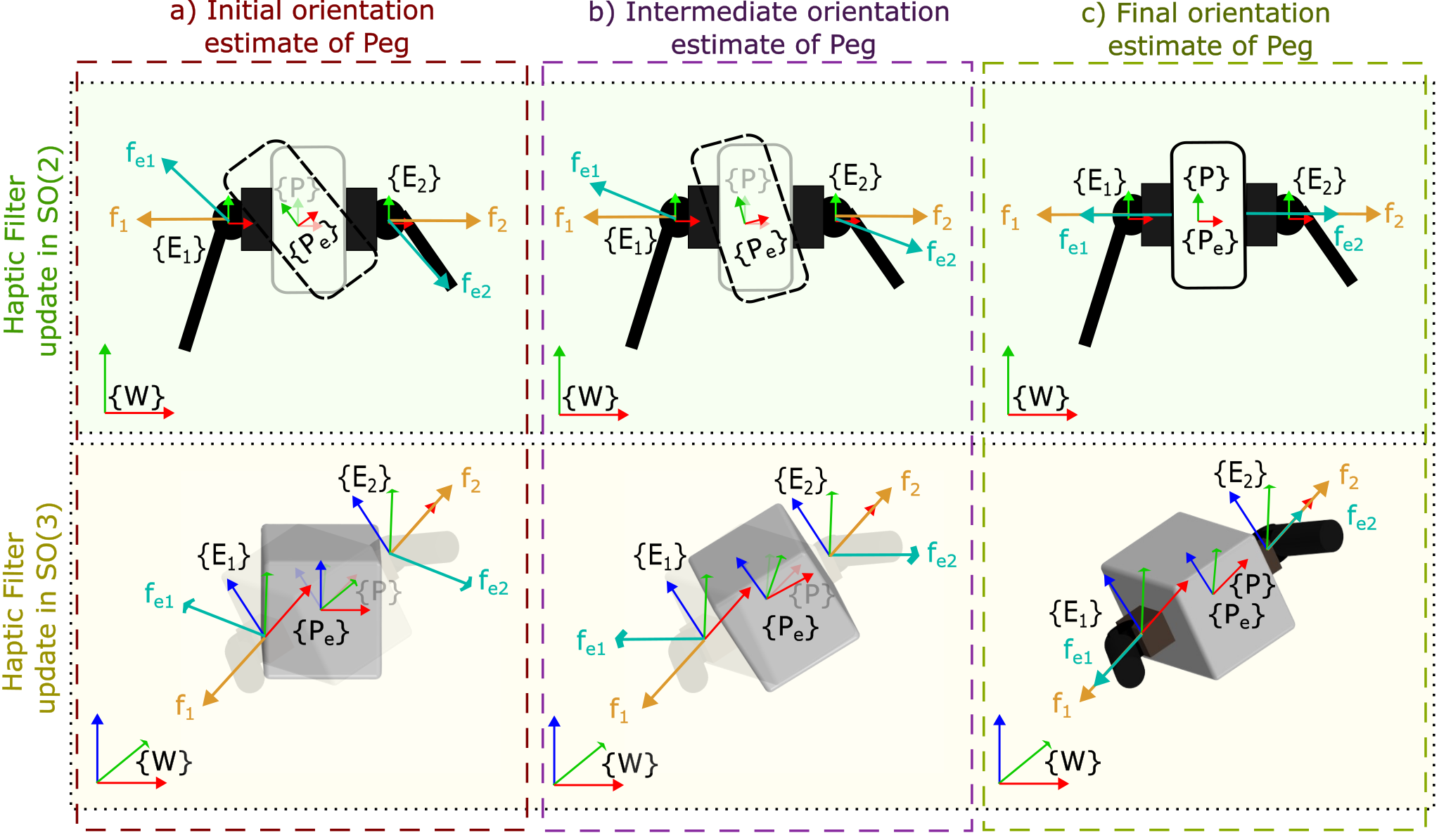}
    \caption{Evolution of the orientation of an object, peg, based on only the haptic information from the force sensors mounted at the end-effectors ($\{E_1\}, \{E_2\}$) of the dual-robotic arm setup. The observer is defined directly on $\SOthree$.}
    \label{fig:haptic_update_2d_3d}
\end{figure}

Several works have proposed the use of haptic filters for different robotic tasks. \citet{Khatib_Kalman_Active_2006} have used Kalman Active Observers for robotic telemanipulation. However, contact forces are inherently non-linear~(\citet{Geo_Frame_Prof_Domenico}). Linearization of a non-linear system cannot guarantee robustness~(\citet{MahonyCompFil}). \citet{particle_filter_pose} presents a pose estimation algorithm using tactile information from the fingertips of a gripper combined with haptic rendering models (HRMs). The CAD file of the gripper's fingertips is used to create a voxmap for HRM. The distance between the object and the voxel is converted into force. This gives the expected tactile responses of the object. A particle filter provides correspondences between the measured sensor information from the real geometry and the expected sensor data from the HRM. As it uses Monte Carlo simulation, this method is computationally expensive. It is well proven that human perception exploits constraints provided by the structure of the scene without reliance on quantitative, pointwise models of the image formation process~(\citet{PENTLAND_AI_Human}). It is then only natural to devise methods that interpret the geometric structure of the visual inputs for tasks such as contact-rich manipulation~(\citet{Gregory_SQ}). One such smooth geometric structure is obtained from superquadric (superellipsoid) modeling, especially useful to capture networks of (convex) rigid bodies with a relatively small number of parameters~(\citet{segRecovSQ}). \citet{Gregory_SQ} have presented a robust method to recover superquadrics from point clouds of convex objects. 

\citet{Petrovskaya} proposed an efficient Bayesian approach termed, Scaling Series, for full 6DOF localization of an object. A touch sensor is attached to the end-effector of a robotic arm. The arm moves along the surface of the stationary object to determine its pose during the sensing phase. It then grasps the object. These methods are clearly not suitable for cases such as robotic insertion tasks where objects can move during the sensing phase.

The kinematics of an object's orientation evolves on a Lie Group~(\citet{bundleObserver_Prof_Ravi}). Exploiting the symmetry of this underlying manifold leads to robustness. \citet{Prof_Domenico_SO3_2006} and~\citet{MahonyCompFil} have presented non-linear complementary filters defined directly on the matrix Lie-group representation of $\SOthree$. These are widely used for estimating the orientation of an object using an Inertial Measurement Unit (IMU) attached to the body frame. The IMU provides measurements of the rate gyro, accelerometer, and magnetometer. These measurements are used to drive the estimator kinematics to the true orientation.

\citet{forniGenAppImpCtrl} have used Virtual Model Control to create energy-based controllers for minimally invasive surgery. These virtual springs and dampers produce controllers that are more interpretable by users. \citet{Prof_Domenico_CDC_2023,Geo_Frame_Prof_Domenico} used these for geometric frameworks to analyze mechanical manipulation under a quasi-static regime. Virtual springs are used to model mechanical interaction. The force generated when two objects come in contact can be estimated using Hooke's law, given a non-linear smooth spring coefficient and the inter-penetration depth. Geometric modeling techniques like superquadrics simplify the calculation of the displacement between two bodies and hence the force using Hooke's law.

In totality, developing non-linear filters defined on $\SOthree$ that can aggregate multiple sensor readings, such as from the camera and force/torque sensor, in a closed loop while exploiting the geometric structure of the objects will lead to high efficiency and robustness. To our knowledge, the existing prior art has not addressed a haptic-based framework defined on $\SOthree$. We propose a haptic-based framework that uses:
\begin{enumerate}
    \item Superquadrics to model objects,
    \item Virtual springs to estimate forces and thus provide haptic mismatch,
    \item Vision sensor to provide orientation measurement, and
    \item Modified non-linear complementary filter defined on $\SOthree$ originally proposed in~(\citet{CDC_Prof_Mahony_2005}).
\end{enumerate}

Fig.~\ref{fig:haptic_update_2d_3d} shows the evolution of an object's, peg (based on peg-in-hole task), orientation based only on the haptic measurement to our observer. Section~\ref{Problem_Statement_and_Setup} describes the details of the problem statement along with the notations used in the paper, Section~\ref{Haptic_based_filtering} mentions the design methodology for the haptic-based filtering framework, and Section~\ref{results} presents the results obtained for different scenarios.

\section{Problem Statement and Setup}
\label{Problem_Statement_and_Setup}
\subsection{Dual-Robotic Arm Setup}
\begin{figure}[!t]
    \begin{center}
    \includegraphics[width=0.7\linewidth]{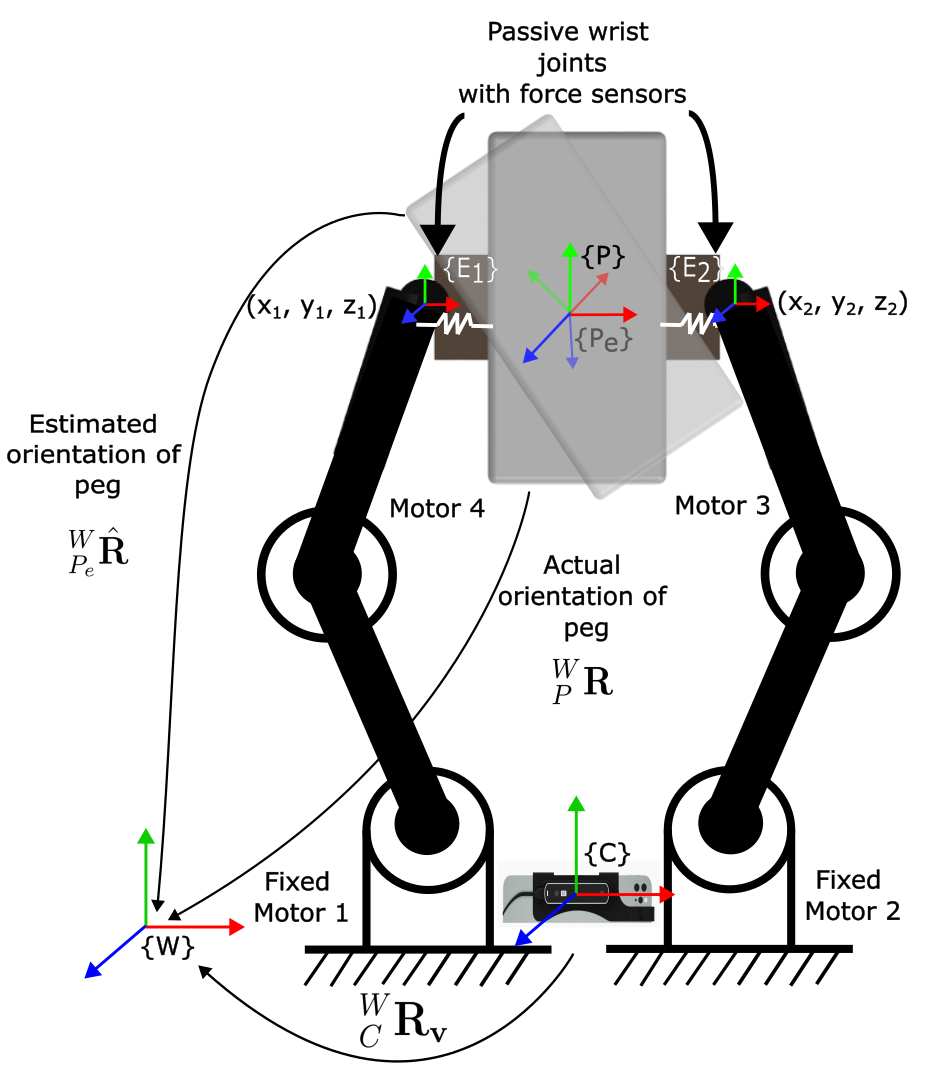}
    \end{center}
    \caption{Dual-robotic arm setup to test the working of designed filter on $\SOthree$.}
    \label{fig:dual_arm_setup_3d}
\end{figure}
Consider a dual-arm setup with four motors mounted at the four revolute joints handling a 
peg as shown in Fig.~\ref{fig:dual_arm_setup_3d}. Although we have considered a 2-link robotic arm for simplicity, our approach remains valid for any n-link manipulator. There is an extensive literature on the dual-arm setup~(\citet{why_dual_arm}). A dual-arm can mimic various tasks, such as a person picking up a box in a warehouse or inserting a component into another. Our proposed filtering framework can also be used on a single robotic arm with a gripper attached; however, this limits the arm's payload capacity.

Force and vision sensors are the measurements to the system. Both end-effectors have force sensors and are mounted on passive revolute joints. The geometry of the peg is fixed and is modeled as a superquadric. The vision sensor measures the orientation of the peg in 3D space. Our analysis begins with the assumption that the dual arms have successfully grasped the peg. This implies that no net force or net torque is acting on the peg.

\begin{figure*}[!t]
    \centering
    \includegraphics[width=0.9\textwidth]{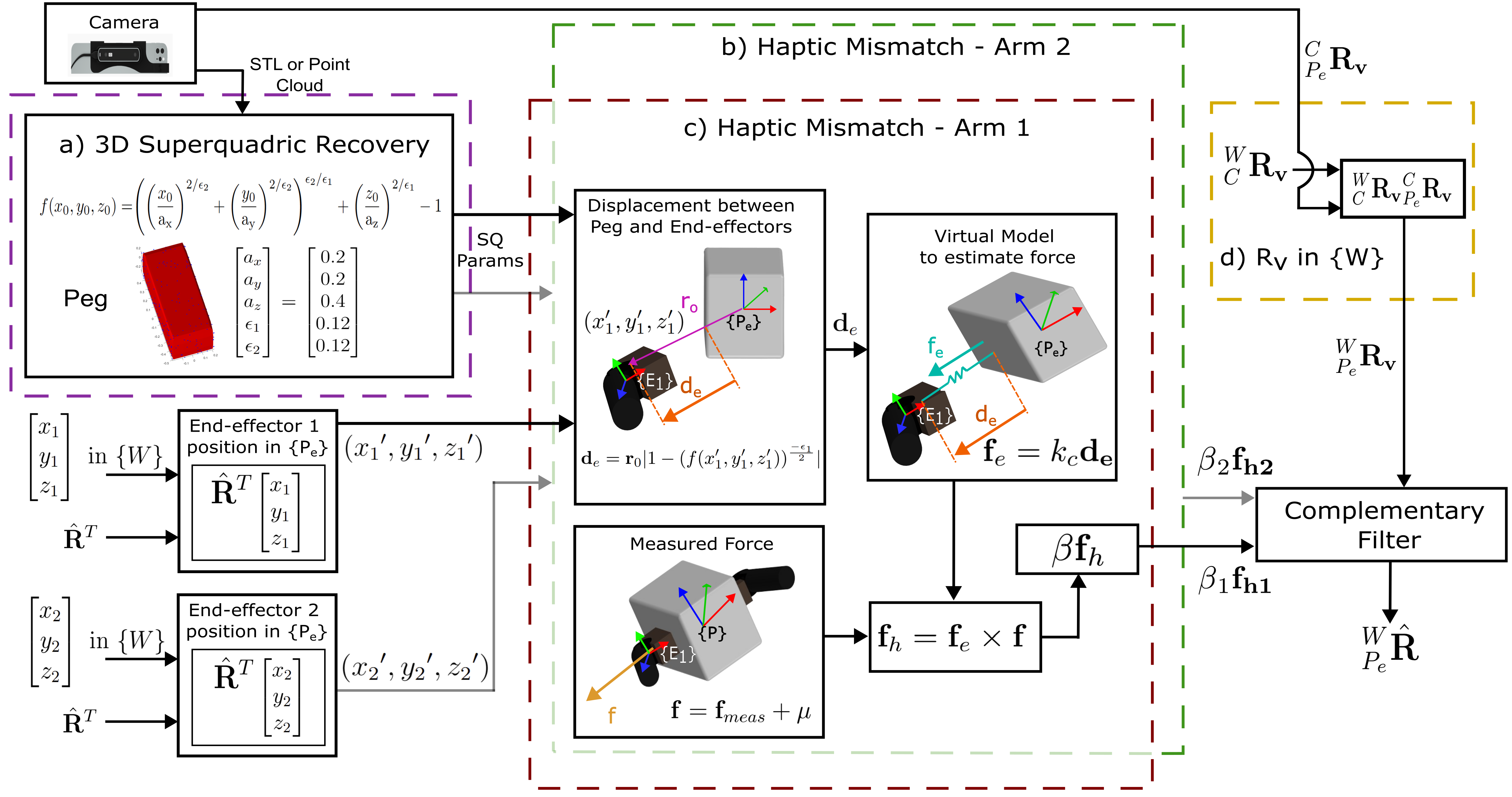}
    \caption{Filtering framework using: a) Superquadrics to model objects, b), c) Virtual springs to estimate forces (haptic measurement) and calculate haptic mismatch, and d) Vision sensor to provide orientation measurement.}
    \label{fig:framework_comp_filter_SO_3_for_haptics}
\end{figure*}

\subsection{Notation}
The following notation is used to denote the frames of reference in the paper-
\begin{itemize}
    \item $\{W\}$ denotes the spatial/fixed frame,
    \item $\{P\}$ denotes the pegs's frame,
    \item $\{P_e\}$ denotes the estimate of peg's frame,
    \item $\{E_1\}$ denotes End-effector frame 1, and
    \item $\{E_2\}$ denotes End-effector frame 2
\end{itemize}
An important point to note here is that the orientation of the end-effectors $\{E_1\}$ and $\{E_2\}$ are not known {\it à priori.} The following notation is used to denote the vectors in the paper-
\begin{itemize}
    \item $\mathbf{r}_0$ denotes the vector from origin of $\{P_e\}$ to origin of $\{E_1\}$ in $\{P_e\}$,
    \item $\mathbf{d}_e$ denotes the radial displacement vector displacement from the surface of the peg (superquadric) to $\{E_1\}$ in $\{P_e\}$,
    \item $\mathbf{f}_e$ denotes the estimated force from the virtual spring in $\{P_e\}$,
    \item $\mathbf{f}$ denotes the measured force from the force/torque sensor in $\{E_1\}$.
    \item $\mathbf{f}_h$ denotes the haptic mismatch in $\{P_e\}$, and
    \item $\boldsymbol{\omega}$ denotes the angular velocity in $\{P_e\}$
\end{itemize}
The rotation matrix, $\mathbf{R}$, and its estimate, $\hat{\mathbf{R}}$, for 
transforming the orientation of the peg to the world-frame, are defined as, 
\begin{align*}
    \mathbf{R} &:= {}^W_{P} {\mathbf{R}} : \{P\} \to \{W\} \nonumber\\
    \hat{\mathbf{R}} &:= {}^W_{P_e} \hat{\mathbf{R}} : \{P_e\} \to \{W\}
\end{align*}
The rotation matrix, $\mathbf{R} \in \SOthree$, denotes orientation of $\{P\}$ relative to $\{W\}$ and is defined as:
\begin{align*}
    \SOthree &= \{ \mathbf{R} \in SL(3, \mathbb{R}) : \mathbf{R}\mathbf{R}^{\top} = I, \det \mathbf{R} = +1 \} 
\end{align*}
and its associated Lie algebra is defined as:
\begin{align*}
    \sothree &= \{ \mathbf{A} \in \mathop{sl(3,\mathbb{R})} : \mathbf{A} + \mathbf{A}^{\top} =0\}
\end{align*}
%
Let \( \boldsymbol{\omega} \in \mathbb{R}^3 \), then we define $\boldsymbol{\omega}_{\times}$ as,

\begin{align*}
    \boldsymbol{\omega}_{\times} =
    \begin{pmatrix}
    0 & -\omega_3 & \omega_2 \\
    \omega_3 & 0 & -\omega_1 \\
    -\omega_2 & \omega_1 & 0
\end{pmatrix} \in \sothree
\end{align*}

For any \( v \in \mathbb{R}^3 \) then \( \boldsymbol{\omega}_{\times} \boldsymbol{v} = \boldsymbol{\omega} \times \boldsymbol{v} \) is the vector cross product. Similarly, the operator vex: $\sothree \to \mathbb{R}^3$ denotes the inverse of the $\boldsymbol{\omega}_{\times}$. $\mathbb{P}_a$ denotes the anti-symmetric projection operator in square matrix space as,
\begin{align*}
    \mathbb{P}_a(\textbf{A}) = \frac{1}{2} (\textbf{A} - \textbf{A}^{\top})
\end{align*}
\section{Haptic-based filtering framework on $\SOthree$}
\label{Haptic_based_filtering}
The overarching philosophy of the filtering framework used in this article is shown in Fig.~\ref{fig:framework_comp_filter_SO_3_for_haptics}. Each part is now discussed in the following sections.
\subsection{Superquadric Modeling and Recovery}
Superquadrics are a family of geometric primitives with a rich shape vocabulary encoded by only five parameters~(\citet{Gregory_SQ}). Superellipsoids are obtained by the spherical product of a pair of superellipses. The general expression of a superquadric~(\citet{segRecovSQ}) (mainly we focus on superellipsoid) is given by:

\begin{align}
    f(x_0, y_0, z_0) =& 
    {\left({\left(\frac{x_0}{\mathrm{a_x}}\right)}^{2/\epsilon_2}+{\left(\frac{y_0}{\mathrm{a_y}}\right)}^{2/\epsilon_2}\right)}^{\epsilon_2/\epsilon_1}  + \nonumber
    \\ & {\left(\frac{z_0}{\mathrm{a_z}}\right)}^{2/\epsilon_1} - 1
    \label{eq:general_sq}
\end{align}
The authors in~(\citet{Gregory_SQ}) have created a MATLAB toolbox to recover superquadrics from point clouds of convex objects. Using this toolbox, we find all the five parameters, $\mathrm{a_x}, \mathrm{a_y}, \mathrm{a_z}, \epsilon_1, \text{ and } \epsilon_2$. This is shown in Fig.~(\getrefnumber{fig:framework_comp_filter_SO_3_for_haptics}a).
\subsection{Haptic Mismatch- Measurement from force sensors}
As mentioned in Section~\ref{Problem_Statement_and_Setup}, our problem statement assumes that the dual arms have successfully grasped the peg. The force sensor gives the actual force measurement, $\mathbf{f}$, and the estimated force, $\mathbf{f}_e$, is calculated using virtual springs connected to the estimated orientation of the peg. There is one spring between each end-effector and the surface of the superquadric. To calculate force using Hooke's law, a radial displacement vector from the surface of the superquadric to the end-effector is required. This is calculated as~(\citet{segRecovSQ}):
\begin{figure}[!t]
    \begin{center}
    \includegraphics[width=\linewidth]{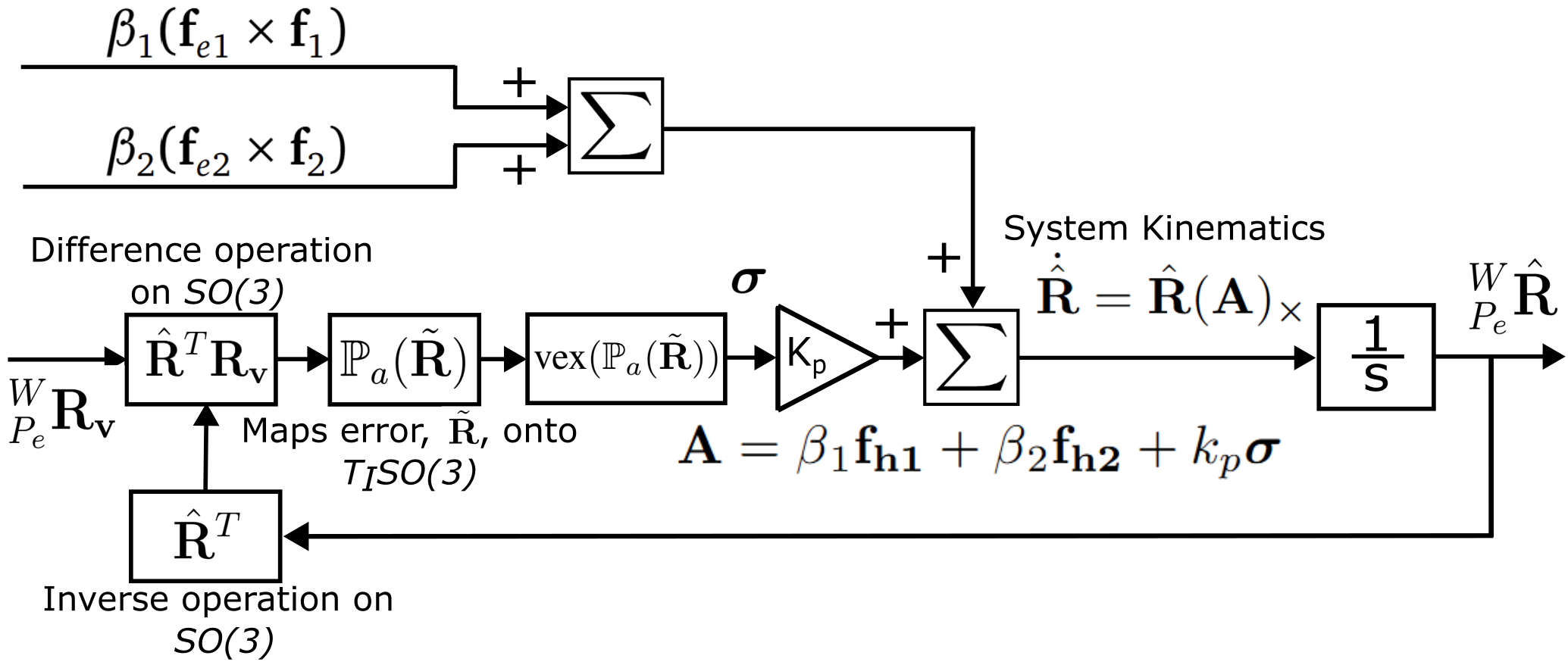}
    \end{center}
    \caption{Modified non-linear complementary filter defined on $\SOthree$}
    \label{fig:modified_comp_filter}
\end{figure}
\begin{align}
    \mathbf{d}_e = \mathbf{r}_0|1 - (f(x_0, y_0, z_0))^{\frac{-\epsilon_1}{2}}|, \ \mathbf{d}_e \in \mathbb{R}^3 \text{ and in } \{P_e\} 
    \label{eq:de_sq}
\end{align}
where, $(x_0, y_0, z_0)$ is the position of the end-effector in $\{P_e\}$ and $\mathbf{r}_0 = [(x_0-0) \ (y_0-0) \ (z_0-0)]^{\top}$ denotes the vector from origin of $\{P_e\}$ to origin of end-effector. As the position of the end-effector is available in $\{W\}$ by using the robotic arm's defined kinematics, the position vector is pre-multiplied by $\hat{\mathbf{R}}^{\top}$ to bring the position vector in $\{P_e\}$. Note that the orientation of the end-effectors is not known a priori.

The spring-force is calculated using Hooke's law as:
\begin{align}
    \mathbf{f}_e = k_c\mathbf{d_e} \in \mathbb{R}^3 \text{ and in } \{P_e\} 
    \label{eq:hooke's_law}
\end{align}
where, $k_c$ is the spring coefficient. $k_c$ can be determined using pre-calibration, depending on the application or the task~(\citet{book_insertion}). The force sensors attached to the end-effectors provide the force measurement, $\mathbf{f_1}$, and $\mathbf{f_2}$ in $\{E_1\}$, and $\{E_2\}$ respectively. The noise in the measurements is denoted by $\mu$. Since we have assumed that the peg is successfully grasped, the orientation of $\{P\}$ remains the same as $\{E_1\}$. Moreover, when $\{P_e\}$ overlaps $\{P\}$, $\mathbf{f}$ can be considered to be in either of the frames. As no net force or torque acts on the peg, the forces measured by both end-effectors will be anti-collinear. Then, the haptic mismatch~(\citet{haptic_mani_harsha}) due to one end-effector is defined as the cross product of $\mathbf{f}$ and $\mathbf{f}_e$, as:

\begin{align}
    \mathbf{f}_h = (\mathbf{f}_e \times \mathbf{f}) \in \mathbb{R}^3
    \label{eq:haptic_mismatch_err}
\end{align}

\begin{figure*}[!t]
    \begin{center}
    \includegraphics[width=\linewidth]{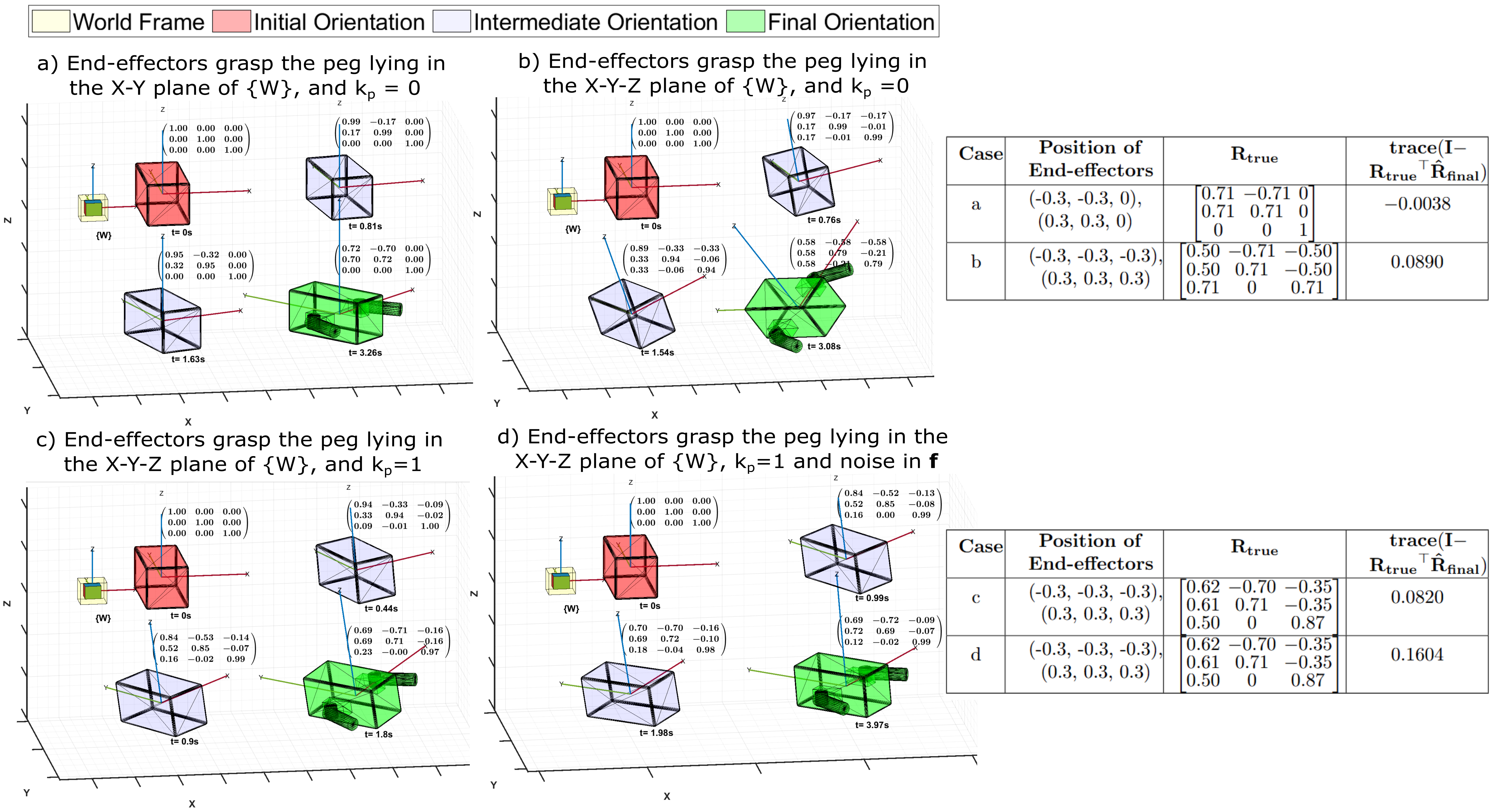}
    \end{center}
    \caption{Evolution of $\hat{\mathbf{R}}$ for different cases based on our designed filtering framework on $\SOthree$.}
    \label{fig:all_result}
\end{figure*}


\noindent $\mathbf{f}_h$ plays the role of a residual-like term in the filter construction. The goal is to drive $\mathbf{f}_h$ to $\textbf{0}$ which will happen when $\mathbf{f}_e = \pm \mathbf{f}$. We exclude the case when $\mathbf{f}_e = - \mathbf{f}$ 
as a measure zero set with highly unlikely occurrence. The change in the value of $\mathbf{f}_e$ as the peg rotates is captured through an admittance coefficient, $\beta$. These steps are shown in Fig.~(\getrefnumber{fig:framework_comp_filter_SO_3_for_haptics}b) and~(\getrefnumber{fig:framework_comp_filter_SO_3_for_haptics}c).
\subsection{Measurement from vision sensor}
The vision sensor measures the orientation of the peg in the camera frame, $\{C\}$. This orientation is defined as:

\begin{align*}
    \mathbf{R}_v &:= {}^C_{P_e} {\mathbf{R}_v} : \{P_e\} \to \{C\}
\end{align*}
As $\{C\}$ is static with respect to $\{W\}$, we obtain the orientation of the peg in the world frame as (shown in Fig.~(\getrefnumber{fig:framework_comp_filter_SO_3_for_haptics}d)):

\begin{align}
    {}^W_{P_e}\mathbf{R}_v = ({}^W_{C} {\mathbf{R}_v}) ({}^C_{P_e} {\mathbf{R}_v}) \in \SOthree
    \label{eq:W_Rv_Pe_vision_measure}
\end{align}

\subsection{Complementary filter on $\SOthree$} 
To achieve our filter design, we take inspiration from the passive complementary filter defined in~\citet{MahonyCompFil} and modify it to suit our problem. The block diagram of the filter is shown in Fig.~\ref{fig:modified_comp_filter}. The rotation kinematics of our filter is given by:
\begin{align}
    \dot{\hat{\mathbf{R}}} = \hat{\mathbf{R}}(\beta{_1}\mathbf{f_{h1}} + \beta{_2}\mathbf{f_{h2}} +  k_p\boldsymbol{\sigma})_{\times}
, \ \hat{\mathbf{R}}(0)=\hat{\mathbf{R}}_0 
\label{eq:kine_our_obs}
\end{align}
%
%
\noindent where $k_p >0$ is a positive gain and $\boldsymbol{\sigma} = \text{vex}(\mathbb{P}_a(\tilde{\mathbf{R}}))$, where the orientation error is:
\begin{align}
    \tilde{\mathbf{R}} =  \hat{\mathbf{R}}^{\top}\mathbf{R}_v \in \SOthree    
    \label{eq:err_obs_design}
\end{align}
and its skew-symmetric component is given as:
\begin{align}
    \mathbb{P}_a(\tilde{\mathbf{R}}) = \frac{1}{2} (\tilde{\mathbf{R}} - \tilde{\mathbf{R}}^{\top}) \in \sothree \text{ and in } \{P_e\}
    \label{eq:Pa_anti_symm_operator_obs}
\end{align}
The term $k_p\boldsymbol{\sigma}$ forms the prediction term of the observer and uses the measurement of the vision sensor. The correction term -  $(\beta{_1}\mathbf{f_{h1}} + \beta{_2}\mathbf{f_{h2}})$ - is based on the force sensors measurement. We analyze each. 
Note that $\mathbb{P}_a(\tilde{\mathbf{R}})$ is invariant under the adjoint map~(\citet{CDC_Prof_Mahony_2005}) as:
\begin{align}
\text{Ad}_{\tilde{\mathbf{R}}} \mathbb{P}_a (\tilde{\mathbf{R}}) = \tilde{\mathbf{R}} \mathbb{P}_a (\tilde{\mathbf{R}}) \tilde{\mathbf{R}}^{\top} = \mathbb{P}_a (\tilde{\mathbf{R}})
\label{eq:Adjoint_pa_invar}
\end{align}
Since we have two end effectors, each with a force sensor, we get two values of haptic mismatch, $\boldsymbol{f_{h1}}$ and $\boldsymbol{f_{h2}}$, as defined in equation~\ref{eq:haptic_mismatch_err} that act as residuals. After multiplication with $\beta$, these are added and we have:
\begin{align}
    \mathbf{A}_{\times} =(\sum_{i=1}^{2}\beta{_i}\mathbf{f_{hi}} \ + \ k_p\boldsymbol{\sigma})_{\times} \in \sothree
    \label{eq:Ax_in_so_3}
\end{align}
The admittance coefficients, $\beta$, for each arm and proportional gain, $k_p$, are the weights of the complementary filter. These weights can be adjusted based on the trust in the measurements from the force and vision sensors.

The continuous-time filter is now discretized for the purpose of implementation. We employ Rodrigues’ formula for integration:
\begin{align}
    \alpha_n &= \mathrm{sinc}(\Delta T \lvert\lvert \mathbf{A}_{n\times}\rvert\rvert) \nonumber \\
    \gamma_n &= \frac{1}{2}(\mathrm{sinc}^{2}\frac{\Delta T \lvert\lvert \mathbf{A}_{n\times}\rvert\rvert}{2}) \nonumber \\
    \hat{\mathbf{R}}_{n+1} &= \hat{\mathbf{R}}_{n}(I + \alpha_n\Delta T \mathbf{A}_{n\times} + \gamma_n (\Delta T\mathbf{A}_{n\times})^2)
    \label{eq:rodrigues_formula}
\end{align}
What remains is checking the stability of our designed filter. The following non-linear stability theorem is similar to the form stated in~(\citet{MahonyCompFil}) and~(\citet{Rodrigues_multimodal_domenico}).
\begin{theorem}
Let $\mathbf{R}(t) \in \SOthree$ represent the orientation of a rigid body. The rotation kinematics of the filter are defined in equation~\ref{eq:kine_our_obs} and measurements given by equations~\ref{eq:haptic_mismatch_err} and~\ref{eq:W_Rv_Pe_vision_measure}. $\mathbf{A}_{\times} \in \sothree$ is defined using equation~\ref{eq:Ax_in_so_3}. Let $\hat{\mathbf{R}}(t)$ denote the solution of equation~\ref{eq:kine_our_obs}. Define error, $\tilde{\mathbf{R}}$, as given in equation~\ref{eq:err_obs_design}. Assume that $\mathbf{A}(t)$ is bounded, absolutely continuous, and that the pair of measurements is asymptotically independent. Define the set, $\mathbb{U}_0 \subseteq \SOthree$ for which the kinematic equation~\ref{eq:kine_our_obs} is unstable. This is given by:

\begin{align}
    \mathbb{U}_0 = \{\tilde{\mathbf{R}} \ | \ \text{tr}(\tilde{\mathbf{R}}) = -1\}
    \label{eq:unstable_set_SO_3}
\end{align}
Then, the error, $\tilde{\mathbf{R}}(t)$ is locally exponentially stable to $I$ and for almost all initial conditions $\tilde{\mathbf{R}}_0 \notin \mathbb{U}_0$, $\hat{\mathbf{R}}(t)$ converges to $\mathbf{R}(t)$.
\end{theorem}

\section{Results}
\label{results}
Using MATLAB's Simulink environment, we have tested the filtering framework shown in Fig.~\ref{fig:framework_comp_filter_SO_3_for_haptics}. The simulations were performed with a fixed time step of 0.01s and ``FixedStepDiscrete" solver. Results are shown in Fig.~\ref{fig:all_result} along with the error in the true and the obtained rotation. The error is calculated using $\text{trace}(\mathbf{I}-\mathbf{R_{true}}^{\top}\mathbf{\hat{R}_{final}})$.

Variables with subscripts 1 and 2 denote each of the end-effectors, respectively. $k_{c1} =1$, $k_{c2} =1$ $\beta_1 =-1$,  $\beta_2 =-1$, $\mathbf{f}_1=
    \begin{bmatrix}
        -1\\
        0\\
        0
    \end{bmatrix}$, and $\mathbf{f}_2=
    \begin{bmatrix}
        1\\
        0\\
        0
    \end{bmatrix}$ for all the cases. We have followed the Z-Y-X rotation order and assumed anti-clockwise rotation about each axis to be positive.

In Fig.~(\getrefnumber{fig:all_result}a), the end-effectors have grasped the peg lying in the X-Y plane of $\{W\}$ and $k_p=0$. Hence,
\begin{align*}
    \begin{bmatrix}
        x_1\\
        y_1\\
        z_1
    \end{bmatrix}&=
    \begin{bmatrix}
        -0.3\\
        -0.3\\
        0
    \end{bmatrix},
    &\begin{bmatrix}
        x_2\\
        y_2\\
        z_2
    \end{bmatrix}=
    \begin{bmatrix}
        0.3\\
        0.3\\
        0
    \end{bmatrix},
    &\hat{\mathbf{R}}_0 =
    \begin{bmatrix}
        1 & 0 & 0\\
        0 & 1 & 0\\
        0 & 0 & 1\\
    \end{bmatrix}
\end{align*}
\normalsize
We can observe in Fig.~(\getrefnumber{fig:all_result}a) that $\hat{\mathbf{R}}$ settles to $\begin{bmatrix}
        0.71 & -0.71 & 0\\
        0.71 & 0.71 & 0\\
        0 & 0 & 1
    \end{bmatrix}$ satisfying the measurements from the force sensors.

In Fig.~(\getrefnumber{fig:all_result}b), the end-effectors have grasped the peg lying in the X-Y-Z plane of $\{W\}$ and $k_p=0$. 
To reach the end-effector positions, the peg is rotated by $45^o$ along the Z axis and then $-45^o$ along the Y axis. Euler angle representation, yaw($\psi$), pitch($\theta$), and roll($\phi$) is also mentioned for easier understanding. Using Fig.~(\getrefnumber{fig:all_result}b) we get,

\begin{align*}
    \mathbf{R_{true}}&=
    \begin{bmatrix}
        0.50 &   -0.71  & -0.50\\
        0.50 &   0.71  & -0.50\\
        0.71 &        0  &  0.71     
    \end{bmatrix} \text{ or }
    \begin{bmatrix}
        \psi_1 \\
        \theta_1\\
        \phi_1
    \end{bmatrix}=
    \begin{bmatrix}
        0.78\\
        -0.78\\
        0
    \end{bmatrix},\\
    \mathbf{\hat{R}_{final}}&=\begin{bmatrix}
        0.58 & -0.58 & -0.58\\
        0.58 & 0.79 & -0.21\\
        0.58 & -0.21 & 0.79
    \end{bmatrix} \text{ or }
    \begin{bmatrix}
        \psi_2 \\
        \theta_2\\
        \phi_2
    \end{bmatrix}=
    \begin{bmatrix}
        0.78\\
        -0.61\\
        -0.26
    \end{bmatrix}    
\end{align*}
\normalsize

The error between the observed and true rotation matrix is simply ${\mathbf{\hat{R}_{final}}}^{\top}\mathbf{R_{true}}$. Converting to Euler angles, we get $(0, -0.17, 0.26)$. This is due to the integration solver settings in MATLAB and the initial condition set.
\begin{figure}[!t]
    \begin{center}
    \includegraphics[width=0.7\linewidth]{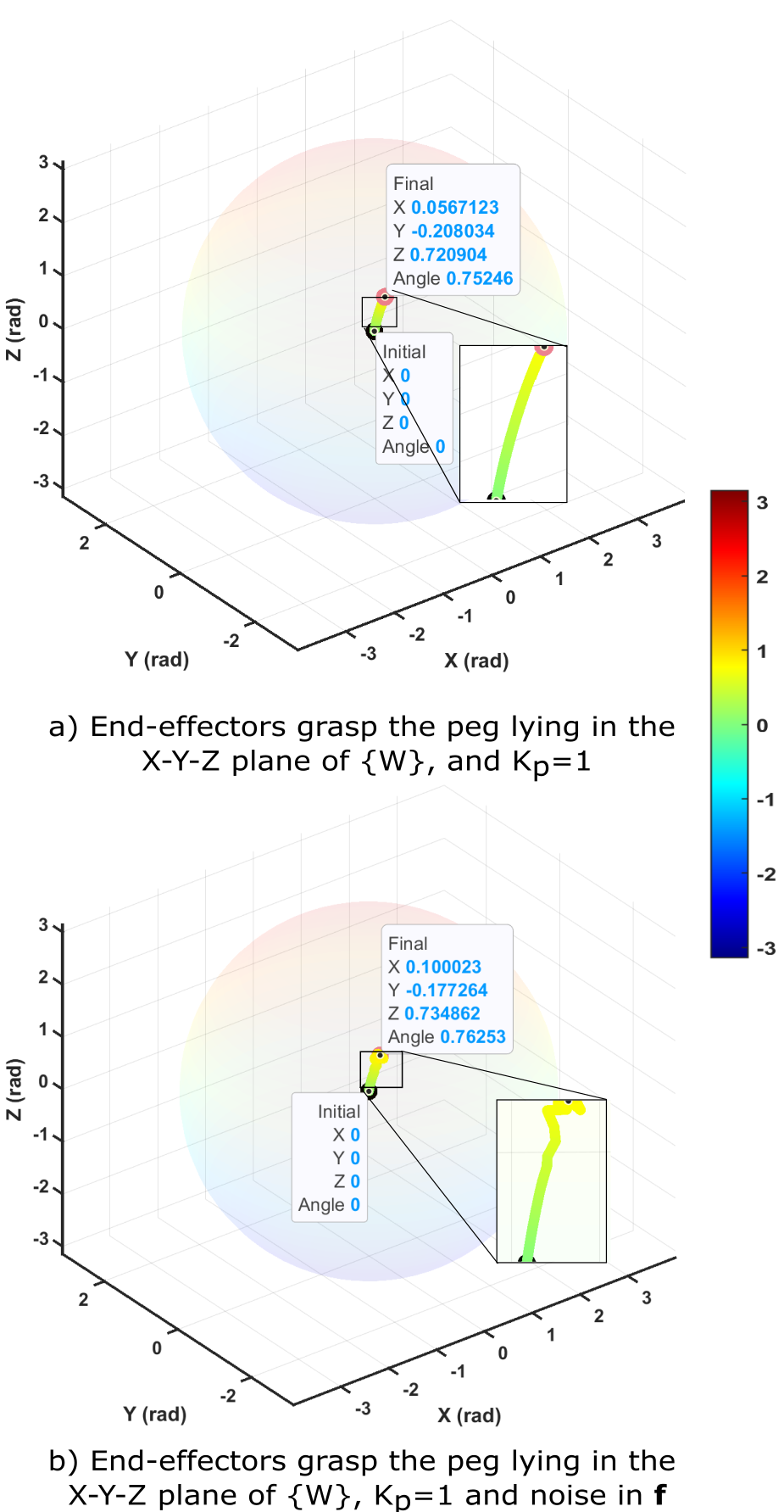}
    \end{center}
    \caption{$\pi$-ball axis-angle representation of $\hat{\mathbf{R}}$ a) Without noise in $\mathbf{f}$, and b) With noise in $\mathbf{f}$.}
    \label{fig:pi_ball_results_3D_noise}
\end{figure}

From Fig.~(\getrefnumber{fig:all_result}c), in addition to the initial conditions for case b, the weight of the measurement from the vision sensor ($k_p$) is considered to be 1. Suppose,
\begin{align*}
\mathbf{R}_v =
    \begin{bmatrix}
        0.71 & -0.71 & 0\\
        0.71 & 0.71 & 0\\
        0 & 0 & 1\\
    \end{bmatrix},
    \begin{bmatrix}
        \psi_3 \\
        \theta_3\\
        \phi_3
    \end{bmatrix}=
    \begin{bmatrix}
        0.78\\
        0\\
        0
    \end{bmatrix}        
\end{align*}
Then we get,

\begin{align*}
    \mathbf{\hat{R}_{final}}&=\begin{bmatrix}
        0.69 & -0.71 & -0.16\\
        0.69 & 0.71 & -0.16\\
        0.23 & 0 & 0.97
    \end{bmatrix},
    \begin{bmatrix}
        \psi_4 \\
        \theta_4\\
        \phi_4
    \end{bmatrix}=
    \begin{bmatrix}
        0.78\\
        -0.23\\
        0
    \end{bmatrix}
\end{align*}

We can observe that the pitch angle $\theta_2 = -0.61$ ($k_p=0$), and $\theta_4=-0.23$ ($k_p=1$) while the $\psi$ angle remains the same. The pitch angle of the peg measured by the vision sensor, $\theta_3=0$, when $k_p=1$. Clearly, $|\theta_2|>|\theta_4|>|\theta_3|$. Fig.~(\getrefnumber{fig:all_result}b) and Fig.~(\getrefnumber{fig:all_result}c) highlight the complementary functionality of our filter, as the fusion of the measurements from the force and the vision sensors results in a final rotation between the ones provided by the individual sensors.




In Fig.~(\getrefnumber{fig:all_result}d), Gaussian noise, with parameters (force sensor 1: $\mu=0,\sigma^2=0.5, \text{sample time}=0.2$, force sensor 2: $\mu=0,\sigma^2=2, \text{sample time}=0.2$) and different seeds for each component, is added to the measured force, $\mathbf{f}$, to analyze the performance of the filter under noisy measurements. For clarity, this is also shown in the $\pi$-ball axis-angle representation of $\hat{\mathbf{R}}$ in Fig.~\ref{fig:pi_ball_results_3D_noise}. The weights can be adjusted for better performance based on the noise characteristics of the sensors.

\begin{figure}[!t]
    \begin{center}
    \includegraphics[width=0.7\linewidth]{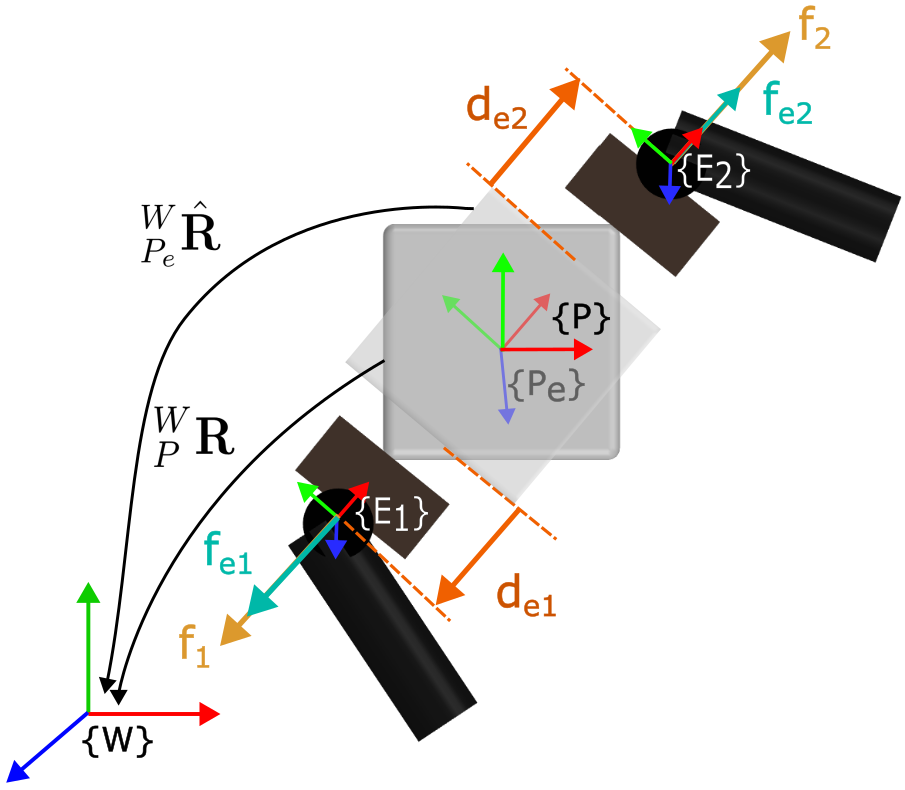}
    \end{center}
    \caption{Representation of a scenario where the end-effectors grasp the peg at the center of the edges.}
    \label{fig:3D_corner_case}
\end{figure}

Consider a possible scenario where the end-effectors grasp the peg at the center of the edges. If $\hat{\mathbf{R}}_0$ is set as shown in Fig.~\ref{fig:3D_corner_case}, the haptic mismatch, $\mathbf{f}_h$, may turn out to be $\mathbf{0}$ even when $\hat{\mathbf{R}} \not= \mathbf{R}$. To get $\mathbf{f}_1$ in $\{P\}$, the rotation, ${}^P_{E1}\mathbf{R}$ is required. An estimate of this can be found using the vision sensor. ${}^P_{E1}\hat{\mathbf{R}}\mathbf{f}_1$ will give the rough measurement in $\{P\}$. The rest of the filtering framework remains as is.

\section{Conclusion and Future Work}
This paper presents a filtering framework defined on $\SOthree$ that incorporates measurements from force and vision sensors to estimate the orientation of a grasped object. Much like human perception, the geometric structure of the object is encoded in the framework using superquadrics. It is validated on a simulated dual-robotic arm setup. Various cases have been detailed that highlight the convergence and robustness of the filter along with its complementary action. 
Future work will focus on online tuning of the admittance coefficient, 6-DOF pose estimation defined on $\mathbb{SE}(3)$, incorporating friction in the analysis and testing on hardware. 

\begin{ack}
The authors would sincerely like to thank Indian Institute of Technology Bombay and Nanyang Technological University for providing the necessary resources to carry out this research. 
\end{ack}


\bibliography{ifacconf}             

@article{SimPLE,
author = {Maria Bauza  and Antonia Bronars  and Yifan Hou  and Ian Taylor  and Nikhil Chavan-Dafle  and Alberto Rodriguez },
title = {SimPLE, a visuotactile method learned in simulation to precisely pick, localize, regrasp, and place objects},
journal = {Science Robotics},
volume = {9},
number = {91},
pages = {eadi8808},
year = {2024},
doi = {10.1126/scirobotics.adi8808},
URL = {https://www.science.org/doi/abs/10.1126/scirobotics.adi8808},
eprint = {https://www.science.org/doi/pdf/10.1126/scirobotics.adi8808}}

@article{JohanssonTactile,
author = {Johansson, Roland and Flanagan, John},
year = {2009},
month = {05},
pages = {345-59},
title = {Coding and use of tactile signals from the fingertips in object manipulation tasks},
volume = {10},
journal = {Nature reviews. Neuroscience},
doi = {10.1038/nrn2621}
}

@INPROCEEDINGS{DecomImiLear,
  author={Xie, Annie and Lee, Lisa and Xiao, Ted and Finn, Chelsea},
  booktitle={2024 IEEE International Conference on Robotics and Automation (ICRA)}, 
  title={Decomposing the Generalization Gap in Imitation Learning for Visual Robotic Manipulation}, 
  year={2024},
  volume={},
  number={},
  pages={3153-3160},
  doi={10.1109/ICRA57147.2024.10611331}}

@INPROCEEDINGS{narang2022factoryfastcontactrobotic,
    AUTHOR    = {Yashraj Narang AND Kier Storey AND Iretiayo Akinola AND Miles Macklin AND Philipp Reist AND Lukasz Wawrzyniak AND Yunrong Guo AND Adam Moravanszky AND Gavriel State AND Michelle Lu AND Ankur Handa AND Dieter Fox}, 
    TITLE     = {{Factory: Fast Contact for Robotic Assembly}}, 
    BOOKTITLE = {Proceedings of Robotics: Science and Systems}, 
    YEAR      = {2022}, 
    ADDRESS   = {New York City, NY, USA}, 
    MONTH     = {June}, 
    DOI       = {10.15607/RSS.2022.XVIII.035} 
}

@ARTICLE{Khatib_Kalman_Active_2006,
  author={Cortesao, R. and Jaeheung Park and Khatib, O.},
  journal={IEEE Transactions on Robotics}, 
  title={Real-time adaptive control for haptic telemanipulation with Kalman active observers}, 
  year={2006},
  volume={22},
  number={5},
  pages={987-999},
  doi={10.1109/TRO.2006.878787}}

@Article{bundleObserver_Prof_Ravi,
title = {A bundle framework for observer design on smooth manifolds with symmetry},
journal = {Journal of Geometric Mechanics},
volume = {13},
number = {2},
pages = {247-271},
year = {2021},
issn = {1941-4889},
doi = {10.3934/jgm.2021015},
url = {https://www.aimsciences.org/article/id/f089b456-a4e9-4204-871a-0109cc7aaae8},
author = {Joshi, Anant A. and Maithripala, D. H. S. and Banavar, Ravi N.}
}

@article{Geo_Frame_Prof_Domenico,
title = {A geometric framework for quasi-static manipulation of a network of elastically connected rigid bodies},
journal = {Applied Mathematical Modelling},
volume = {143},
pages = {116003},
year = {2025},
issn = {0307-904X},
doi = {https://doi.org/10.1016/j.apm.2025.116003},
url = {https://www.sciencedirect.com/science/article/pii/S0307904X25000782},
author = {Domenico Campolo and Franco Cardin}
}

@ARTICLE{MahonyCompFil,
  author={Mahony, Robert and Hamel, Tarek and Pflimlin, Jean-Michel},
  journal={IEEE Transactions on Automatic Control}, 
  title={Nonlinear Complementary Filters on the Special Orthogonal Group}, 
  year={2008},
  volume={53},
  number={5},
  pages={1203-1218},
  doi={10.1109/TAC.2008.923738}}

@article{particle_filter_pose,
author = {Ding, Yitao and Bonse, Julian and Andre, Robert and Thomas, Ulrike},
title = {In-Hand Grasping Pose Estimation Using Particle Filters in Combination with Haptic Rendering Models},
journal = {International Journal of Humanoid Robotics},
volume = {15},
number = {01},
pages = {1850002},
year = {2018},
doi = {10.1142/S0219843618500020},
URL = { 
        https://doi.org/10.1142/S0219843618500020},
eprint = { 
        https://doi.org/10.1142/S0219843618500020}
}

@Article{Lin_Prof_Domenico_Insertion_Task,
AUTHOR = {Yang, Lin and Ariffin, Mohammad Zaidi and Lou, Baichuan and Lv, Chen and Campolo, Domenico},
TITLE = {A Planning Framework for Robotic Insertion Tasks via Hydroelastic Contact Model},
JOURNAL = {Machines},
VOLUME = {11},
YEAR = {2023},
NUMBER = {7},
ARTICLE-NUMBER = {741},
URL = {https://www.mdpi.com/2075-1702/11/7/741},
ISSN = {2075-1702},
DOI = {10.3390/machines11070741}
}

@ARTICLE{Petrovskaya,
  author={Petrovskaya, Anna and Khatib, Oussama},
  journal={IEEE Transactions on Robotics}, 
  title={Global Localization of Objects via Touch}, 
  year={2011},
  volume={27},
  number={3},
  pages={569-585},
  doi={10.1109/TRO.2011.2138450}}

@article{PENTLAND_AI_Human,
title = {Perceptual organization and the representation of natural form},
journal = {Artificial Intelligence},
volume = {28},
number = {3},
pages = {293-331},
year = {1986},
issn = {0004-3702},
doi = {https://doi.org/10.1016/0004-3702(86)90052-4},
url = {https://www.sciencedirect.com/science/article/pii/0004370286900524},
author = {Alex P. Pentland}
}

@INPROCEEDINGS{Gregory_SQ,
  author={Liu, Weixiao and Wu, Yuwei and Ruan, Sipu and Chirikjian, Gregory S.},
  booktitle={2022 IEEE/CVF Conference on Computer Vision and Pattern Recognition (CVPR)}, 
  title={Robust and Accurate Superquadric Recovery: a Probabilistic Approach}, 
  year={2022},
  volume={},
  number={},
  pages={2666-2675},
  doi={10.1109/CVPR52688.2022.00270}}

@book{segRecovSQ,
  title={Segmentation and Recovery of Superquadrics},
  author={Jaklic, Ales and Leonardis, Ales and Solina, Franc},
  year={2000},
  publisher={Springer}
}

@article{forniGenAppImpCtrl,
title = {A Generalized Approach to Impedance Control Design for Robotic Minimally Invasive Surgery},
journal = {IFAC-PapersOnLine},
volume = {56},
number = {2},
pages = {8548-8555},
year = {2023},
note = {22nd IFAC World Congress},
issn = {2405-8963},
doi = {https://doi.org/10.1016/j.ifacol.2023.10.015},
url = {https://www.sciencedirect.com/science/article/pii/S2405896323003531},
author = {Daniel Larby and Fulvio Forni}
}

@ARTICLE{why_dual_arm,
  author={Merlo, Elena and Lagomarsino, Marta and Ajoudani, Arash},
  journal={IEEE Robotics and Automation Letters}, 
  title={Information-Theoretic Detection of Bimanual Interactions for Dual-Arm Robot Plan Generation}, 
  year={2025},
  volume={},
  number={},
  pages={1-8},
  doi={10.1109/LRA.2025.3552216}}

@Article{dual_arm_Harsha_Prof_Domenico,
AUTHOR = {Turlapati, Sri Harsha and Campolo, Domenico},
TITLE = {Towards Haptic-Based Dual-Arm Manipulation},
JOURNAL = {Sensors},
VOLUME = {23},
YEAR = {2023},
NUMBER = {1},
ARTICLE-NUMBER = {376},
URL = {https://www.mdpi.com/1424-8220/23/1/376},
PubMedID = {36616974},
ISSN = {1424-8220},
DOI = {10.3390/s23010376}
}

@Article{haptic_mani_harsha,
AUTHOR = {Turlapati, Sri Harsha and Accoto, Dino and Campolo, Domenico},
TITLE = {Haptic Manipulation of 3D Scans for Geometric Feature Enhancement},
JOURNAL = {Sensors},
VOLUME = {21},
YEAR = {2021},
NUMBER = {8},
ARTICLE-NUMBER = {2716},
URL = {https://www.mdpi.com/1424-8220/21/8/2716},
PubMedID = {33921508},
ISSN = {1424-8220},
DOI = {10.3390/s21082716}
}

@INPROCEEDINGS{CDC_Prof_Mahony_2005,
  author={Mahony, R. and Hamel, T. and Pflimlin, J.-M.},
  booktitle={Proceedings of the 44th IEEE Conference on Decision and Control}, 
  title={Complementary filter design on the special orthogonal group SO(3)}, 
  year={2005},
  volume={},
  number={},
  pages={1477-1484},
  doi={10.1109/CDC.2005.1582367}}

@INPROCEEDINGS{Prof_Domenico_CDC_2023,
  author={Campolo, Domenico and Cardin, Franco},
  booktitle={2023 62nd IEEE Conference on Decision and Control (CDC)}, 
  title={Quasi-Static Mechanical Manipulation as an Optimal Process}, 
  year={2023},
  volume={},
  number={},
  pages={4753-4758},
  doi={10.1109/CDC49753.2023.10383481}}

@INPROCEEDINGS{Prof_Domenico_SO3_2006,
  author={Campolo, Domenico and Keller, Flavio and Guglielmelli, Eugenio},
  booktitle={2006 IEEE/RSJ International Conference on Intelligent Robots and Systems}, 
  title={Inertial/Magnetic Sensors Based Orientation Tracking on the Group of Rigid Body Rotations with Application to Wearable Devices}, 
  year={2006},
  volume={},
  number={},
  pages={4762-4767},
  doi={10.1109/IROS.2006.282346}}

@INPROCEEDINGS{Rodrigues_multimodal_domenico,
  author={Campolo, Domenico and Schenato, Luca and Pi, Lijuan and Deng, Xinyan and Guglielmelli, Eugenio},
  booktitle={2008 IEEE/RSJ International Conference on Intelligent Robots and Systems}, 
  title={Multimodal sensor fusion for attitude estimation of micromechanical flying insects: A geometric approach}, 
  year={2008},
  volume={},
  number={},
  pages={3859-3864},
  doi={10.1109/IROS.2008.4650812}}

@ARTICLE{book_insertion,
  author={Yang, Lin and Turlapati, Sri Harsha and Lv, Chen and Campolo, Domenico},
  journal={IEEE Robotics and Automation Letters}, 
  title={Planning for Quasi-Static Manipulation Tasks via an Intrinsic Haptic Metric: A Book Insertion Case Study}, 
  year={2025},
  volume={10},
  number={6},
  pages={6111-6118},
  doi={10.1109/LRA.2025.3564707}}
                                                   







\end{document}